\documentclass[11pt]{article}

\usepackage[final]{acl}

\usepackage{times}
\usepackage{latexsym}
\usepackage{booktabs}
\usepackage{multirow}
\usepackage[T1]{fontenc}

\usepackage[utf8]{inputenc}

\usepackage{microtype}

\usepackage{inconsolata}

\usepackage{graphicx}
\usepackage{amsmath}
\usepackage{xcolor}
\usepackage{colortbl}
\usepackage{hhline}
\usepackage{makecell}   
\usepackage{xspace}
\usepackage{arydshln}
\usepackage{amssymb}

\usepackage{amsmath,amsfonts,bm}

\definecolor{MySalmonrgb}{rgb}{0.9216,0.3608,0.3608}

\def\eqref#1{equation~\ref{#1}}

\def\1{\bm{1}}

\DeclareMathAlphabet{\mathsfit}{\encodingdefault}{\sfdefault}{m}{sl}
\SetMathAlphabet{\mathsfit}{bold}{\encodingdefault}{\sfdefault}{bx}{n}

\newcommand{\KL}{D_{\mathrm{KL}}}

\definecolor{stdblue}{HTML}{0047AB}    
\definecolor{darkorange}{HTML}{CC5500} 

\newcommand{\epslow}{\textcolor{stdblue}{\epsilon_{\text{lo}}}}
\newcommand{\epshigh}{\textcolor{darkorange}{\epsilon_{\text{hi}}}}
\newcommand{\epshimax}{\textcolor{darkorange}{\epsilon_{\text{hi}}^{\max}}}
\newcommand{\epslowbold}{\textcolor{stdblue}{\boldsymbol{\epsilon}_{\text{\textbf{lo}}}}}
\newcommand{\epshighbold}{\textcolor{darkorange}{\boldsymbol{\epsilon}_{\text{\textbf{hi}}}}}

\newcommand{\Ai}{A_i}                                  
\newcommand{\rhoi}{\rho_i}                             
\newcommand{\rhoistar}{\rho_i^{*}}                     

\newcommand{\yi}{y_i}                                  
\newcommand{\seqlen}{|\yi|}                            

\newcommand{\pinew}{\pi_\theta}                        
\newcommand{\piold}{\pi_{\theta_{\text{old}}}}         
\newcommand{\pinewy}{\pinew(\yi \mid x)}               
\newcommand{\pioldy}{\piold(\yi \mid x)}               

\newcommand{\sigspo}{s_i(\theta)}                      

\newcommand{\Dkl}[2]{\KL(#1 \,\|\, #2)}
\newcommand{\KLrev}{\Dkl{\pinew}{\piold}}              
\newcommand{\KLfwd}{\Dkl{\piold}{\pinew}}              
\newcommand{\KLrevcond}{\Dkl{\pinew(\cdot \mid x)}{\piold(\cdot \mid x)}}
\newcommand{\KLfwdcond}{\Dkl{\piold(\cdot \mid x)}{\pinew(\cdot \mid x)}}

\newcommand{\Exp}[1]{\exp\!\left(#1\right)}            
\newcommand{\clipf}[3]{\mathrm{clip}\!\left(#1,\, #2,\, #3\right)}  
\newcommand{\Expect}[1]{\mathbb{E}\!\left[#1\right]}   

\newcommand{\Lgapo}{\mathcal{L}^{\text{GAPO}}}

\newcommand{\Lsurr}{L_{\piold}(\pinew)}                

\newcommand{\Jtheta}{J(\theta)}                        

\newcommand{\rbar}{\bar{r}}                            
\newcommand{\Amax}{A_{\max}}                           

\newcommand{\ee}[1]{\mathrm{e}{#1}}                    
\newcommand{\hExpecti}[1]{\widehat{\mathbb{E}}_{i}\!\left[#1\right]}

\usepackage[framemethod=TikZ]{mdframed}
\usepackage{caption}
\definecolor{commentred}{HTML}{B22222}

\mdfdefinestyle{thinkstyle}{%
  linewidth=0.4pt,
  linecolor=black!85,
  backgroundcolor=gray!6,
  roundcorner=4pt,
  innertopmargin=1.2ex,
  innerbottommargin=1.0ex,
  innerleftmargin=1.5ex,
  innerrightmargin=1.5ex,
  frametitleaboveskip=4pt,
  frametitlebelowskip=4pt,
  frametitlefont=\bfseries\normalsize\color{white},
  frametitlebackgroundcolor=black!85,
  frametitlerule=false,
  skipabove=6pt, skipbelow=6pt,
}
\newenvironment{thinkbox}[1]{%
  \begin{mdframed}[style=thinkstyle, frametitle={#1}]%
}{%
  \end{mdframed}%
}
\newenvironment{problembox}[1]{%
  \begin{mdframed}[style=thinkstyle, frametitle={#1}]%
}{%
  \end{mdframed}%
}
\definecolor{myred}{RGB}{220,20,60}
\definecolor{myblue}{RGB}{30,144,255}
\definecolor{mygreen}{RGB}{34,139,34}
\definecolor{mypurple}{RGB}{138,43,226}

\newcommand{\camera}[1]{#1}

\newcommand{\daggernote}[1]{%
  \begingroup
  \renewcommand{\thefootnote}{}
  \renewcommand{\theHfootnote}{daggernote}
  \footnote{\dag\,#1}
  \addtocounter{footnote}{-1}
  \endgroup
} 

\title{Group Adaptive Clipping Policy Optimization}

\author{
  \textbf{Sheng Jia}\textsuperscript{1,2},
  \textbf{Xiao Wang}\textsuperscript{2},
  \textbf{Shiva Prasad Kasiviswanathan}\textsuperscript{2\dag},
  \textbf{Rein Houthooft}\textsuperscript{2\dag}
\\
  \textsuperscript{1}University of Toronto,
  \textsuperscript{2}Amazon
\\
  \small \texttt{\{\href{mailto:shengama@amazon.com}{shengama},
  \href{mailto:waxao@amazon.com}{waxao},
  \href{mailto:kasivisw@amazon.com}{kasivisw},
  \href{mailto:reinrh@amazon.com}{reinrh}\}@amazon.com}
}

\begin{document}
\maketitle
\daggernote{Equal advising.}

\begin{abstract}

Group relative policy optimization for reinforcement learning with verifiable rewards (RLVR) typically uses a fixed importance-sampling (IS) ratio clipping boundary across all rollouts. We identify a key limitation: rare correct rollouts on harder problems and abundant correct rollouts on easier problems are clipped at comparable rates, despite contributing very different learning signals. Rollouts with low group success exhibit larger IS ratios and carry stronger gradient signal for exploration and solving new problems, yet are disproportionately suppressed by fixed clipping. 

To address this, we propose \textit{Group Adaptive Clipping Policy Optimization (GAPO)}, a plug-in modification to GRPO methods that adapts the clipping boundary to the rollout advantage. GAPO is motivated by a reverse-KL trust-region perspective, which suggests that rollouts with larger learning signal should receive proportionally greater update headroom. GAPO requires no reward shaping and preserves the standard PPO/GSPO surrogate while adapting only the clipping threshold. Across Qwen and Llama models, GAPO consistently improves both Pass@1 and Pass@k over fixed clipping and advantage-shaping baselines on math reasoning and coding benchmarks where the pass rates by the base model are relatively low\footnote{Code available:\,\url{https://github.com/Sheng-J/GAPO}. Correspondence to:\,\href{mailto:shengama@amazon.com}{shengama@amazon.com}}.
\end{abstract}

\section{Introduction}
\label{sec:intro}
\begin{figure}[t]
  \includegraphics[width=\columnwidth]{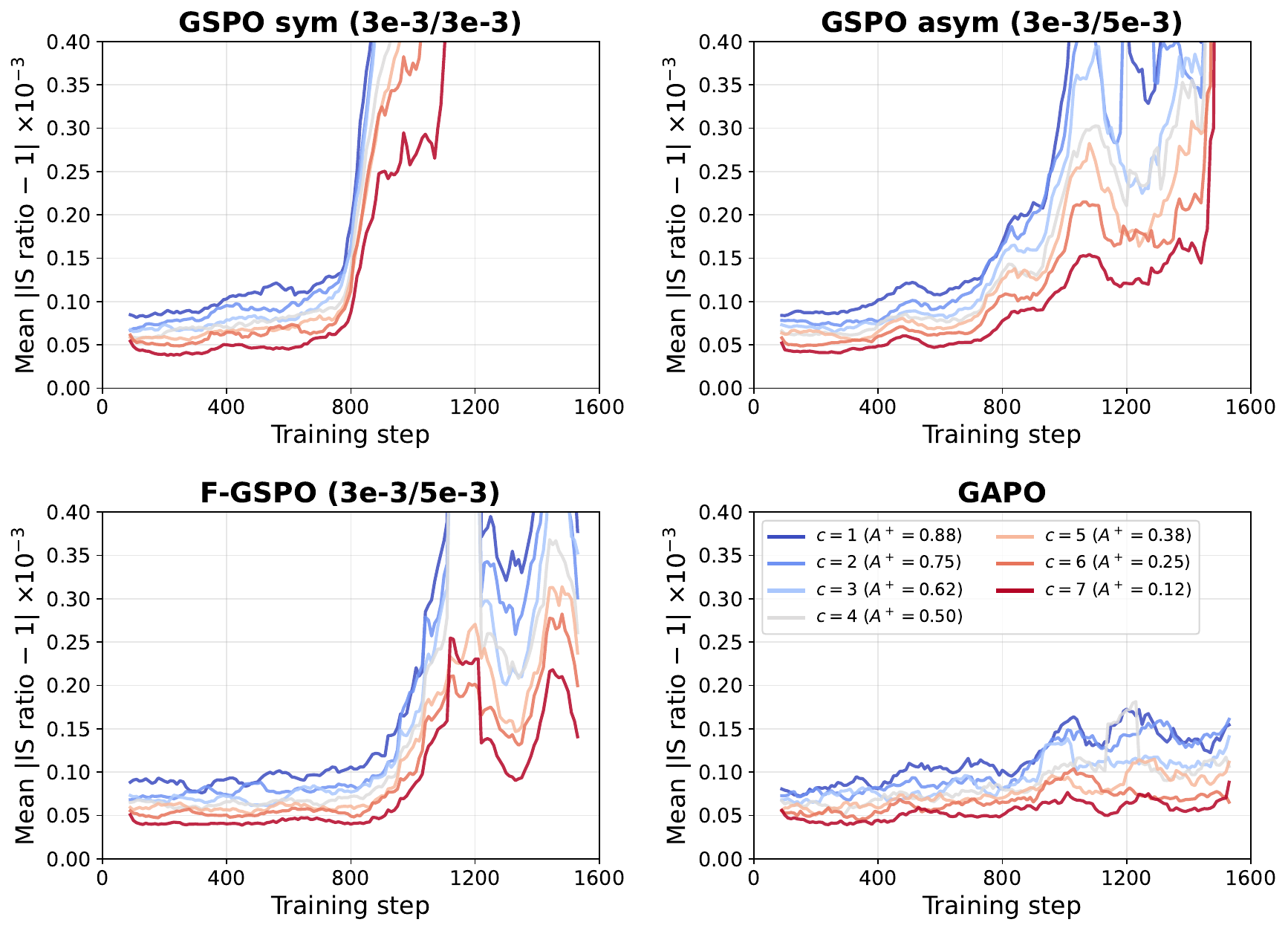}
\caption{\textbf{Mean deviation of the importance-sampling (IS) ratio over RLVR training, binned by correctness count $c$ within a group.} Two patterns hold across algorithms: (1) the IS ratio is consistently higher for scarcely correct rollouts (small $c$, fewer correct solutions per group), and (2) for these rollouts the ratio continues to grow at a rate equal to or exceeding that of rollouts on easier problems (large $c$). Together these patterns motivate adapting the clip boundary to question difficulty: under a fixed boundary, the rare high-advantage rollouts on hard problems hit the clip first and are suppressed disproportionately early.} 
  \label{fig:IS_ratio_mean_dev}
\end{figure}

Reinforcement learning with verifiable rewards (RLVR) trains language models using PPO-style clipped policy optimization~\citep{schulman2017proximal}, where a fixed importance-sampling (IS) ratio boundary approximates a trust region by suppressing updates once the policy moves too far from its previous iterate. In group-relative RLVR methods such as GRPO~\citep{shao2024deepseekmathpushinglimitsmathematical}, GSPO~\citep{zheng2025group}, and DAPO~\citep{yu2025dapo_NEURIPS2025_a4277440}, rollouts are assigned advantages relative to other samples from the same prompt. Specifically, in a group of $k$ rollouts with $c$ correct solutions, correct rollouts receive advantage $\Ai = (k-c)/k$, creating a structured spectrum ranging from scarce, high-advantage correct rollouts (low $c$) to abundant, low-advantage correct rollouts (high $c$).

A fixed clipping boundary treats this spectrum uniformly, despite substantial differences in learning signal. We observe that before clipping activates, IS ratios grow proportionally to advantage: scarce correct rollouts on harder problems move away from the reference policy substantially faster than abundant correct rollouts on easier problems (Figure~\ref{fig:IS_ratio_mean_dev}). Yet once clipping begins, both are truncated at comparable rates under a uniform boundary. This creates a mismatch: scarce correct rollouts, which provide valuable signal for exploration and solving new problems, are clipped too aggressively, while abundant redundant rollouts are granted equal update headroom despite contributing less to improvement.

A natural response might be to widen the clipping boundary uniformly, as explored in DAPO~\citep{yu2025dapo_NEURIPS2025_a4277440}. However, this fails to address the core issue: the problem is not the absolute width of the clipping boundary, but its uniformity across rollouts with fundamentally different advantages. What must change is the \emph{relative} clip width across different levels of group success.


Under a reverse-KL trust-region perspective, the optimal policy update at a single prompt allocates probability to responses proportionally to their advantage, yielding a trust-region-optimal importance-sampling (IS) ratio that scales exponentially with advantage~\citep{ppolambda}. This suggests a simple principle for clipping: rollouts with larger learning signal should receive proportionally greater update headroom. Pushing an IS ratio substantially beyond this optimum either violates the trust region or over-allocates probability mass to a single response at the expense of alternatives, making the optimal ratio a natural guide for adaptive clipping.

We leverage this observation to introduce \textit{Group Adaptive Policy Optimization (GAPO)}, a simple plug-in modification to group-relative policy optimization that adapts the per-rollout clip threshold to rollout advantage. In RLVR, binary rewards induce only a small number of discrete positive advantage levels, allowing the adaptive clipping threshold to be computed directly from the group success statistic $c$. GAPO preserves the standard PPO/GSPO surrogate objective and modifies only the clipping boundary. Unlike reward- or advantage-shaping approaches, GAPO continues to optimize pass@1 directly and does not exhibit the pass@1 drift that can arise from shaping objectives based on difficulty~\citep{plyusov2026f}, inference-time pass@$k$~\citep{walder2025passkoptimization_NEURIPS2025_df8a1a63, chen2025passktrainingbytedance, tang2025optimizinginferencetimeobjectives}, or diversity~\citep{li2025jointlydarling}. We make the following contributions:


\begin{itemize}
    \item We identify a \emph{clipping asymmetry} in RLVR: under uniform clip boundaries, scarce correct rollouts (low $c$, high advantage) are clipped at rates comparable to abundant correct rollouts, despite carrying substantially stronger learning signal (Figure \ref{fig:clip_frac}). We show that a reverse-KL trust-region perspective (Equation \ref{eq:reverse_constrained}, \ref{eq:lagrangian_perprompt}) naturally motivates \emph{advantage-dependent clipping}, yielding a trust-region-optimal IS ratio that scales exponentially with advantage and reduces to a practical closed-form clipping rule in RLVR through the group statistic $c$ (Equation \ref{eq:optimal_policy}).
    
    \item We introduce Group Adaptive Clipping Policy Optimization (GAPO), a simple plug-in modification to PPO/GSPO that adapts only the clipping boundary while preserving the standard surrogate objective and direct optimization of pass@1 (Equation \ref{eq:gapo}).
    
    \item We demonstrate consistent improvements in pass@1 and pass@$k$ across Qwen2.5-Math-1.5B, Llama-3.2-3B-Instruct, and DeepSeek-R1-Distill-Qwen-1.5B on mathematical reasoning and \camera{code generation} benchmarks (Table~\ref{tab:gspo_variants}, \ref{tab:coding_appendix}, \ref{tab:all_models}, Figure \ref{fig:valid_pass}), while maintaining strong correlation between IS ratio and advantage throughout training (Figure \ref{fig:IS_ratio_corr}).  Figure~\ref{fig:ruleout_confounds_IS_adv_correlation_ckpt600_branch} shows a checkpoint intervention experiment to rule out common confounds.
\end{itemize}

\begin{figure*}[t]
    \includegraphics[width=1.0\linewidth]{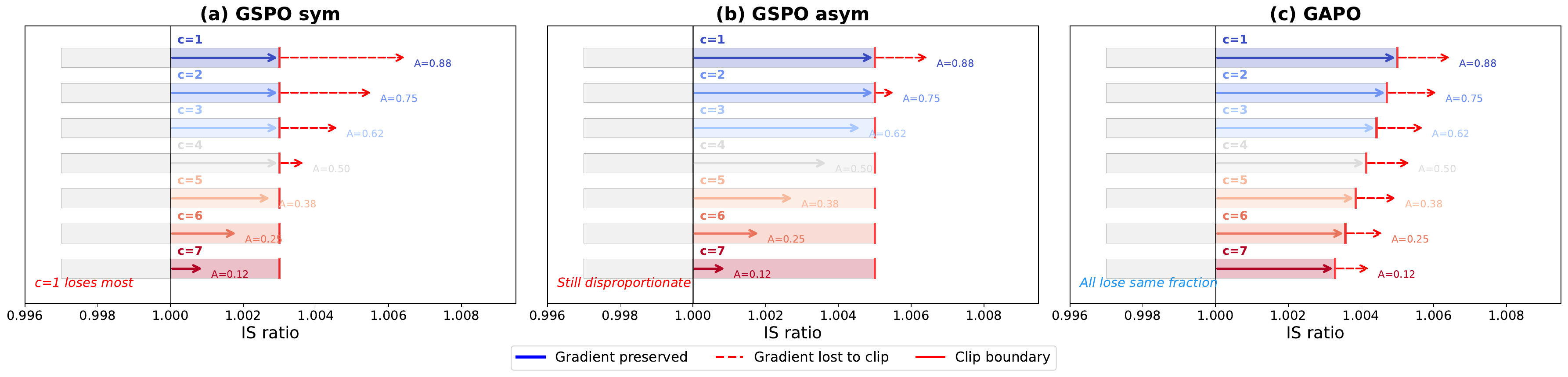} \hfill
    \caption{\textbf{(a) Symmetric clipping (b) Asymmetric Clipping (c) Adaptive Clipping.} Each row represents a correct rollout with group statistic $c$ and advantage $\Ai$. Solid arrows show preserved gradient; dashed red arrows show gradient lost to IS surpassing clipping thresholds. (a) GSPO symmetric: a uniform boundary clips scarce correct rollouts disproportionately. (b) GSPO asymmetric: wider upper boundary but still uniform across $c$. 
    (c) GAPO: per-$c$ adaptive boundary~\eqref{eq:gapo} keeps gradient updates proportional to advantage~\eqref{eq:optimal_policy} even as clipping starts firing.}
\label{fig:adap_vs_fix_clip}
\end{figure*}
\section{Preliminaries}
\label{sec:prelim}

\subsection{RL with Verifiable Rewards}
\label{sec:prelim-rlvr}

We consider reinforcement learning with verifiable rewards (RLVR) for
language model post-training. Given a prompt $x$ drawn from a dataset
$\mathcal{D}$, the policy $\pinew$ generates a complete response
$y \sim \pinew(\cdot \mid x)$, which receives a binary verifiable
reward $R(x, y) \in \{0, 1\}$ from an automated verifier. The training
objective is the expected reward
\begin{equation}
\begin{split}
    \Jtheta = \;\; &\Expect{R(x, y)}, \\
    &x \sim \mathcal{D},\; y \sim \pinew(\cdot \mid x).
\end{split}
\label{eq:rlvr_objective}
\end{equation}
In group-relative policy optimization with $k$ rollouts per prompt
$\{y_i\}_{i=1}^k \sim \piold(\cdot \mid x)$, a correct rollout $i$ in
a group with $c$ correct has advantage
\begin{equation}
\Ai = r_i - \rbar = 1 - \frac{c}{k} = \frac{k - c}{k},
\label{eq:advantage}
\end{equation}
where $r_i \in \{0, 1\}$.
Scarce correct rollouts ($c{=}1$) receive large advantage $(k{-}1)/k$,
while abundant correct rollouts ($c{=}k{-}1$) receive small advantage
$1/k$.  In this work, we do not perform advantage normalization for unbiased advantage estimate  \citep{liu2025understanding_r1zero}.  

Standard PPO-style clipping applies a fixed trust region
$[1-\epsilon,\,1+\epsilon]$ uniformly to all rollouts. We observe that
this disproportionately suppresses high-advantage (scarce) rollouts,
which push their importance ratio past the clip boundary fastest.

\subsection{Trust Region Policy Optimization}
\label{sec:prelim-trpo}

\paragraph{Trust regions and reverse KL.}
We build on the policy-improvement guarantee of \citet{schulman2015trust}.
In the RLVR setting the bound takes the form
\begin{multline}
\Jtheta \geq \Lsurr \\
- C \cdot \max_{x}\, \KLfwdcond,
\label{eq:trpo_bound}
\end{multline}
where the surrogate, written in importance-sampling form, is
\begin{equation}
\begin{aligned}
\Lsurr
&\;=\;
\hExpecti{\frac{\pinew(y_i \mid x)}{\piold(y_i \mid x)}\, \Ai} \\
&\;=\;
\hExpecti{\rhoi \, \Ai},
\end{aligned}
\label{eq:surrogate_is}
\end{equation}
with $\hExpecti{\cdot}$ denoting the empirical expectation over a
batch of rollouts $\{(x, y_i)\}$ drawn from $\piold(\cdot \mid x)$,
and $C$ depending only on the reward range. The bound follows from
Pinsker's inequality applied to the worst-case total-variation
distance between policies; maximising its right-hand side guarantees
monotonic improvement of $\Jtheta$ at each update.

Equation \ref{eq:trpo_bound} is stated with the forward direction $\KLfwd$,
but the same guarantee holds with the reverse
direction~\citep{ppolambda}. The reason is symmetry: because
$D_{\mathrm{TV}}(P, Q) = D_{\mathrm{TV}}(Q, P)$, Pinsker's inequality
bounds the same quantity from either side, so $\KLrev$ may replace
$\KLfwd$ in \eqref{eq:trpo_bound} without weakening the guarantee. The
constrained optimization we solve is therefore
\begin{equation}
\begin{aligned}
\max_{\theta}\;\;     & \hExpecti{\rhoi \, \Ai} \\
\text{s.t.}\;\;       & \hExpecti{\KLrevcond} \;\leq\; \delta,
\end{aligned}
\label{eq:reverse_constrained}
\end{equation}
where the empirical expectation over the batch replaces the worst-case
maximum for tractability, as in PPO~\citep{schulman2017proximal}.

 \begin{figure*}[t]
     \includegraphics[width=1.0\linewidth]{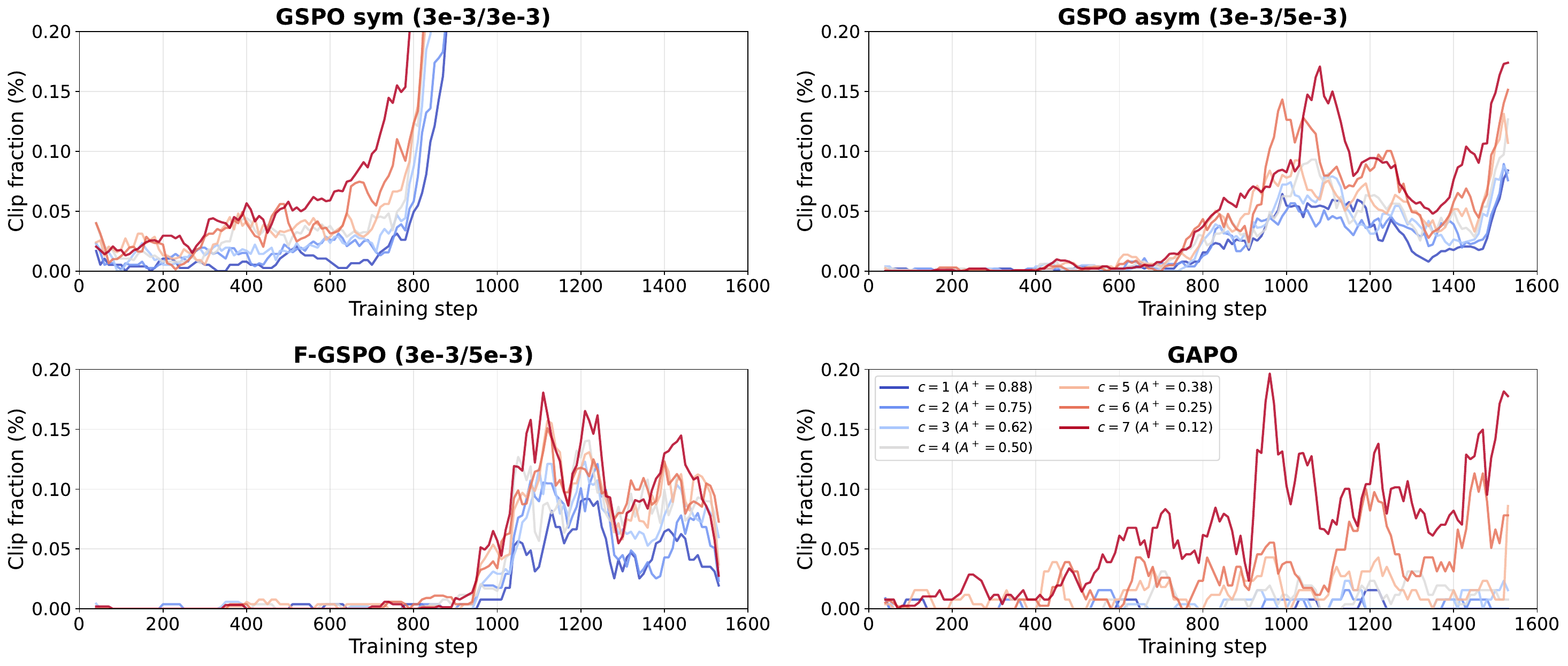} \hfill
     \caption{\textbf{Token clip fraction over training of Qwen2.5-Math-1.5B, by group correctness $c$.} Fraction of tokens clipped \emph{within} rollouts at each $c$. Under fixed clipping (GSPO sym / asym), scarce correct rollouts on harder questions (low $c$, larger advantage $\Ai$~\eqref{eq:advantage}) are clipped at rates comparable to abundant correct rollouts (high $c$). GAPO clips high-$c$ rollouts more and low-$c$ rollouts less, retaining gradient on the rare correct rollouts most valuable for exploration.}
     \label{fig:clip_frac}
 \end{figure*}
\section{Group Adaptive Policy Optimization }
\label{sec:gapo}

\subsection{Per-Prompt Trust Region Optimization}
\label{sec:gapo-derivation}
Following \citet{ppolambda}, we consider local policy optimization at a state $s$, which is a prompt $x$ in our case. This makes the batch-level constraint in equation \ref{eq:reverse_constrained} per-prompt, exact only for single-prompt batches (see Limitations). The
Lagrangian of maximizing \eqref{eq:reverse_constrained} given $x$ is
\begin{equation}
\begin{aligned}
\mathcal{L}_x = {} & \sum_{y} \pinew(y|x)\, A(x,y) \\
                   & - \lambda \KLrevcond.
\end{aligned}
\label{eq:lagrangian_perprompt}
\end{equation}
By solving the
following Euler--Lagrange equation \citep{gelfand2000calculus}, 
\begin{equation}
\begin{aligned}
\frac{\partial \mathcal{L}_x}{\partial \pinew(y|x)}
\!&=\! A(x,y) \!-\! \lambda\!\left(\log \frac{\pinew(y|x)}{\piold(y|x)} \!+\! 1\right) \\
&= 0.
\end{aligned}
\label{eq:euler_lagrange}
\end{equation}
we obtain the stationary point for the target policy $\pi^{*}_{\theta}(y|x) \;\propto\; \piold(y|x)\,\Exp{\frac{A(x,y)}{\lambda}}.$
At each rollout $i$ for prompt $x$, the optimal IS ratio is
\begin{equation}
\rhoistar = \frac{\pi^{*}_{\theta}(y_i|x)}{\piold(y_i|x) }\propto \exp(\Ai/\lambda).
\label{eq:optimal_policy}
\end{equation}
\paragraph{Interpretation on the optimal IS ratio.}
$\rhoistar$ describes how much the probability of generating
response $y_i$ should be pushed up for prompt $x$ to maximize
the expected return on prompt $x$. Pushing $\rhoi$ past
$\rhoistar$ means either the trust region constraint is
violated, or probability of generating some other response
$y_j$ is not pushed according to \eqref{eq:optimal_policy},
lowering the expected return under this formulation. The clip
should therefore fire at $\rhoistar$, giving
\begin{equation}
\epshigh(i) \;\geq\; \rhoistar - 1
\;=\; \Exp{\frac{\Ai}{\lambda}} - 1.
\label{eq:clip_accommodation}
\end{equation}
GSPO uses clip widths $\epsilon \sim 10^{-3}$, and GAPO inherits this scale, so $\epshigh \sim 10^{-3}$. Inverting $\epshigh = \exp(\Ai/\lambda) - 1$ gives $|\Ai/\lambda| = \log(1+\epshigh) \approx 10^{-3}$, hence $\lambda \sim \Amax/10^{-3} \sim 10^3 \gg \Amax$ (with $\Amax = (k-1)/k \leq 1$). The linearization $\exp(\Ai/\lambda) \approx 1 + \Ai/\lambda$ then gives
\begin{equation}
\epshigh(i) \;\approx\; \frac{\Ai}{\lambda} \;=\; \frac{k - c_i}{k\lambda}.
\label{eq:linearized}
\end{equation}
\camera{The linearization is only for presentation cleanliness. At the sequence-IS level it is essentially exact.  
At the token-IS scale $\epshimax{=}0.28$, it is conservative (smaller $\epsilon_{\text{hi}}$ than exact, so still within the trust region), preserves the monotonic ordering in $c$, and keeps the relative error on $\Ai/\lambda$ bounded ($2\text{--}13\%$).}
\subsection{GAPO Adaptive Clip Formula}
\label{sec:gapo-formula}

Normalising to interpolate between $\epslow$ (minimum, for $c=k$) and
$\epshimax$ (maximum, for $c=1$), we obtain the GAPO adaptive upper
clip:
\begin{equation}
\boxed{\;\epshigh(c) \;=\; \epslow + (\epshimax - \epslow) \cdot \frac{k - c}{k - 1}\;}
\label{eq:gapo}
\end{equation}
with boundaries
$\epshigh(c{=}1) = \epshimax, \epshigh(c{=}k) = \epslow.$
Scarce correct rollouts ($c{=}1$) receive maximum headroom, while
abundant correct rollouts ($c \to k$) are constrained to the minimum.
Incorrect rollouts always use $\epslow$ (the upper clip is irrelevant
for negative advantage).  \camera{For rollouts with $\Ai<0$ where $\epslow$ can become active, the update scales with $|A|=\frac{c}{k}$, which is small on hard problems (low $c$), precisely where correct rollouts are scarce.  Adaptive $\epslow$ matters more for the trust-region of penalizing incorrect rollouts on easy problems (high $c$). It affects neither the positive signal nor the clipping bias that prevents exploration \citep{yu2025dapo_NEURIPS2025_a4277440}, so we fix $\epslow$. }

Combined with GSPO's sequence-level importance ratio $\sigspo$, the
full GAPO objective is
\begin{multline}
\Lgapo = \mathbb{E}\Bigl[\max \Bigl(-\Ai \cdot \sigspo, \\
-\Ai \cdot \clipf{\sigspo}{1 - \epslow}{1 + \epshigh(c_i)}\Bigr)\Bigr],
\label{eq:gapo_objective}
\end{multline}
where the sequence-level ratio is the geometric mean of per-token
ratios:
\begin{equation}
\sigspo
\;=\; \left(\frac{\pinewy}{\pioldy}\right)^{1/\seqlen}.
\label{eq:gspo_ratio}
\end{equation}
Both GAPO's adaptive clip and GSPO's sequence-level ratio operate at
the rollout level, and the clip threshold directly gates the
sequence-level ratio derived from the same per-rollout trust region
argument.

%
%

\section{Experiments \& Results}
\label{sec:experiments&results}
\begin{figure*}[t]
    \includegraphics[width=0.48\linewidth]{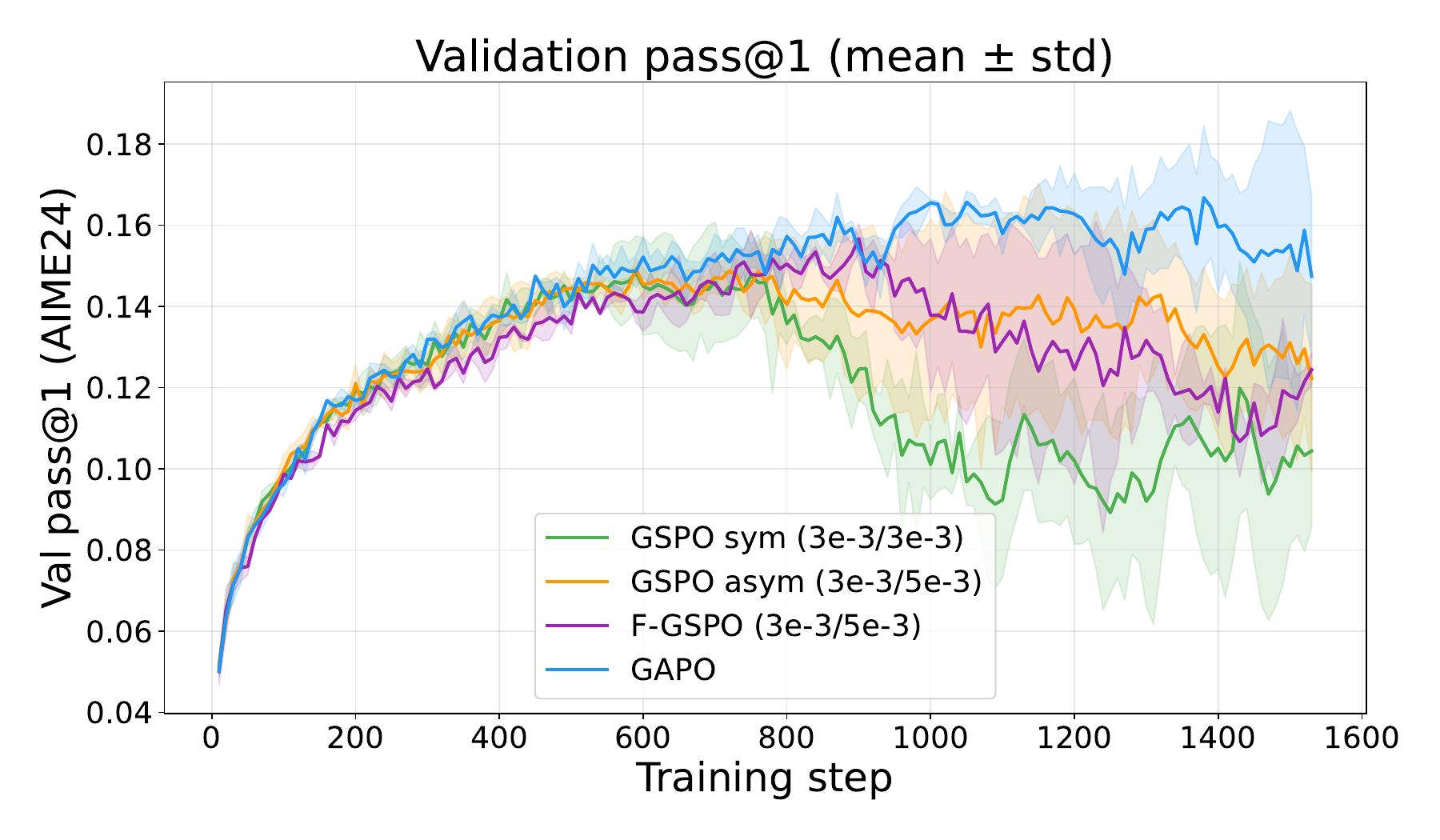} \hfill
    \includegraphics[width=0.48\linewidth]{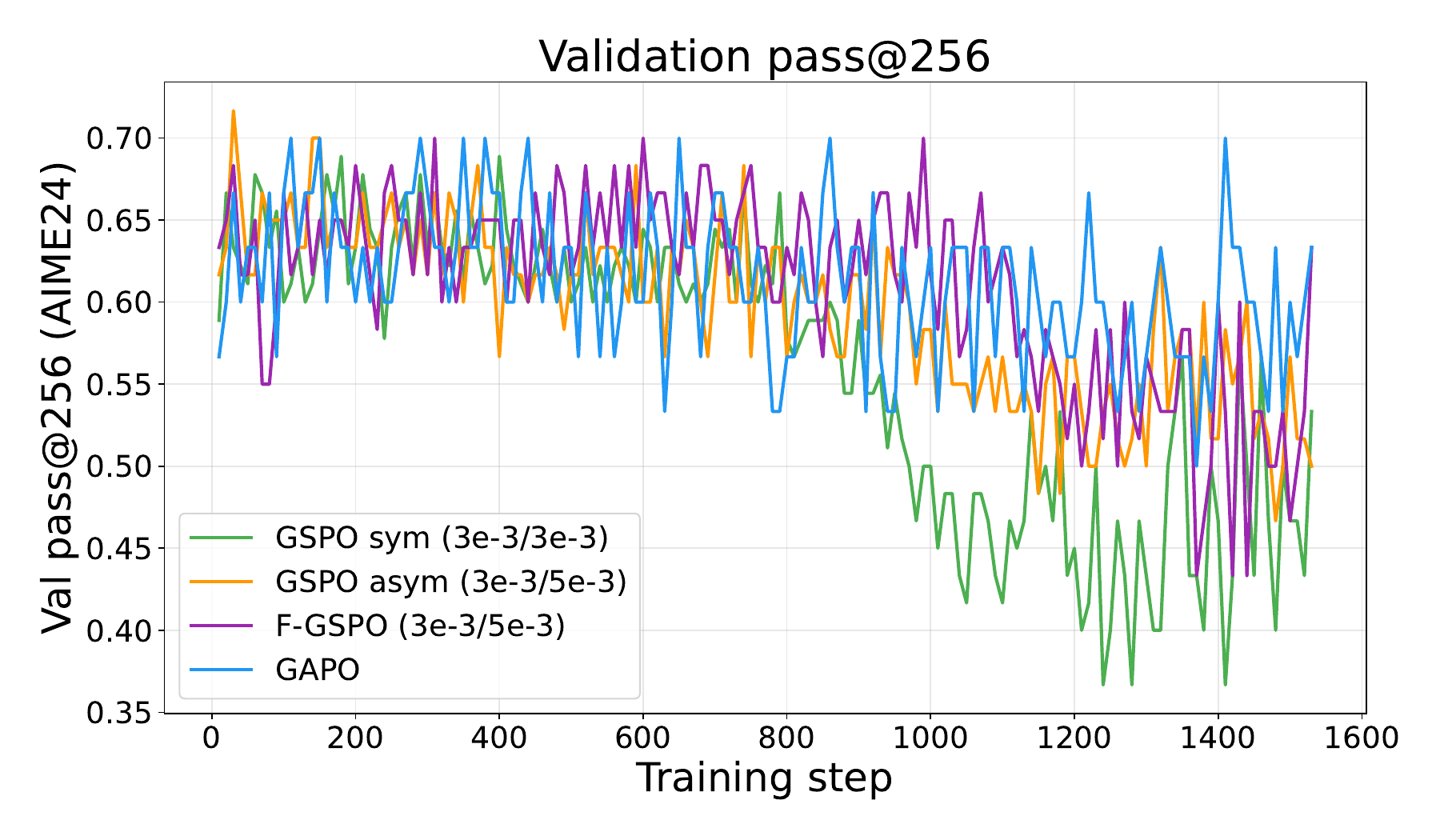} \hfill
\caption{\textbf{Validation pass@1 and pass@256 on AIME24 over training.} Qwen2.5-Math-1.5B under four RLVR algorithms (mean ± std across seeds). GAPO sustains higher pass@1 and pass@256 in late training, while fixed-clip baselines (especially symmetric GSPO) show pass@256 collapse after step $\sim 1000$. The diversity loss adaptive clipping is designed to avoid by retaining gradient on low-$c$ rollouts (Figures~\ref{fig:clip_frac}, \ref{fig:IS_ratio_corr}).}    \label{fig:valid_pass}
\end{figure*}
We first compare the training dynamics of GAPO on Qwen2.5-Math-1.5B with three fixed clipping baselines.  Specifically, we are interested in (i) how clip fraction within rollouts at each $c$ differs between fixed and adaptive clipping (Figure~\ref{fig:clip_frac}), (ii) which clipping schedule maintains a high correlation between the empirical IS ratio and advantage (Figure~\ref{fig:IS_ratio_corr}), and (iii) whether this correlation translates into higher pass@1 and pass@$k$ (Figure~\ref{fig:valid_pass}).  We then benchmark GAPO against representative RLVR methods on two base models without SFT (Qwen2.5-Math-1.5B, Llama-3.2-3B-Instruct) in Table \ref{tab:all_models} and one reasoning-distilled base model (DeepSeek-R1-Distill-Qwen-1.5B) in Table \ref{tab:gspo_variants}, \ref{tab:coding_appendix}.
\subsection{Experiment Setup}
\paragraph{Models and datasets.}
We train and evaluate three base models spanning two model
families and two pretraining regimes: Qwen2.5-Math-1.5B
(math-pretrained) \citep{yang2024qwen2}, Llama-3.2-3B-Instruct (general-purpose
instruction-tuned) \citep{grattafiori2024llama}, and DeepSeek-R1-Distill-Qwen-1.5B
(reasoning-distilled). We include this post-SFT model, since their RLVR usually has a different training dynamic, and we aim to show GAPO works for all settings. Math RL runs train on the DeepScaleR dataset with 39{,}202 samples after filtering duplicates  \citep{tan2026deepscaler}.  \camera{For the RL runs with code generation data in Table~\ref{tab:coding_appendix}, we use 24{,}269 samples from DeepCoder~\citep{luo2025deepcoder} after filtering overlong prompts: 16{,}238 from SYNTHETIC-1~\citep{2025synthetic1}, 7{,}432 from TACO~\citep{likaixin2024taco-verified}, and 599 from LiveCodeBench 2023-5-1 to 2024-7-31~\citep{jain2025livecodebench}.}
\paragraph{Training details.}
Our main method builds on GSPO's sequence-level importance ratio
$\sigspo$~\eqref{eq:gspo_ratio}, with clip boundaries
$(\epslow,\,\epshimax)=(3\ee{-3},\, 5\ee{-3})$ for Qwen2.5-Math-1.5B experiments and $(\epslow, \epshimax)=(7\ee{-5},\, 3\ee{-4})$ \camera{for DS-R1-distilled models, as distilled models have small IS drifts}.
For token-IS, we use clip boundaries
$\epslow = 0.2$ and $\epshimax = 0.28$.  \camera{For RLVR with code generation data, we follow DeepCoder recipe \citep{luo2025deepcoder} and also use token-IS.}
We use group
size $k = 8$. Loss is aggregated at the token level (sum of
per-token losses divided by total token count). We do not
normalise the advantage by
$\mathrm{std}(\{R(q, o_1), \dots, R(q, o_k)\})$, following
Dr.GRPO~\citep{liu2025understanding_r1zero}; this std
normalisation introduces a question-level difficulty bias,
upweighting groups whose rewards are nearly all 1 or 0. \camera{Up-weighting groups with many rewards=1 can counteract updates with rare correct ones.}  Full hyperparameters are listed in Table~\ref{tab:hyperparams},\,\ref{tab:hyperparams_coding}.

\paragraph{Baselines.}
We compare against fixed-clip RLVR methods at matched clip
widths: GRPO~\citep{shao2024deepseekmathpushinglimitsmathematical},
Dr.GRPO~\citep{liu2025understanding_r1zero}, GSPO with symmetric
and asymmetric clip ranges~\citep{zheng2025group}, and the
focal-shaping variants F-GRPO~\citep{plyusov2026f} and F-GSPO. These apply advantage shaping $\widetilde{A}_i = (1 - c/k)^{\gamma}\,\Ai$ with $\gamma \geq 0$.
This selection spans the standard group-relative baselines and
the two main alternative directions for adapting the surrogate
(advantage shaping and asymmetric clipping) for exploration. The symmetric clipping $(\epslow{=}3e{-}3, \epshimax{=}3e{-}3)$ baseline is also chosen to show that GAPO's improvement comes not from clipping tighter, but from the relative difference in clip width across rollouts with different group success rates.  We also evaluate a token-IS variant of GAPO against token-IS baselines. Among non-adaptive methods, GSPO's sequence-level IS is the strongest baseline in our setting, so we adopt it as the base for both GAPO and the advantage-shaping baseline F-GSPO.

\paragraph{Evaluation.}
We evaluate on standard mathematical reasoning benchmarks
(AIME24, AIME25 \citep{ye2025aimepreview}, AMC, MATH500 \citep{hendrycks2021measuringmath500}, Minerva \citep{minervamath_lewkowycz2022solving}, OlympiadBench \citep{he2024olympiadbench}) \camera{and code generation benchmarks (LiveCodeBench \citep{jain2025livecodebench}  , HumanEval+ \citep{liu2023yourHumanEvalPlus})}, and we report
IFEval \citep{zhou2023instructionIFEVALBENCH} as an out-of-domain probe for instruction-following
retention. For Qwen2.5-Math-1.5B \citep{yang2024qwen2} and Llama-3.2-3B-Instruct \citep{grattafiori2024llama}, we
report pass@1 and pass@256 with temperature $1.0$, top-$p=1$,
$n = 256$ samples per prompt, and $T_{\max}$ of $3092$ and
$8192$ respectively. For DeepSeek-R1-Distill-Qwen-1.5B we
report pass@1 and pass@16 with temperature $0.6$, top-$p$ $1$,
$n = 16$, and $T_{\max} = 24576$ to accommodate longer rollouts.  

\subsection{Adaptive clipping preserves the IS--advantage correlation}
\label{sec:exp-dynamics}
\begin{figure}[t]
  \includegraphics[width=\columnwidth]{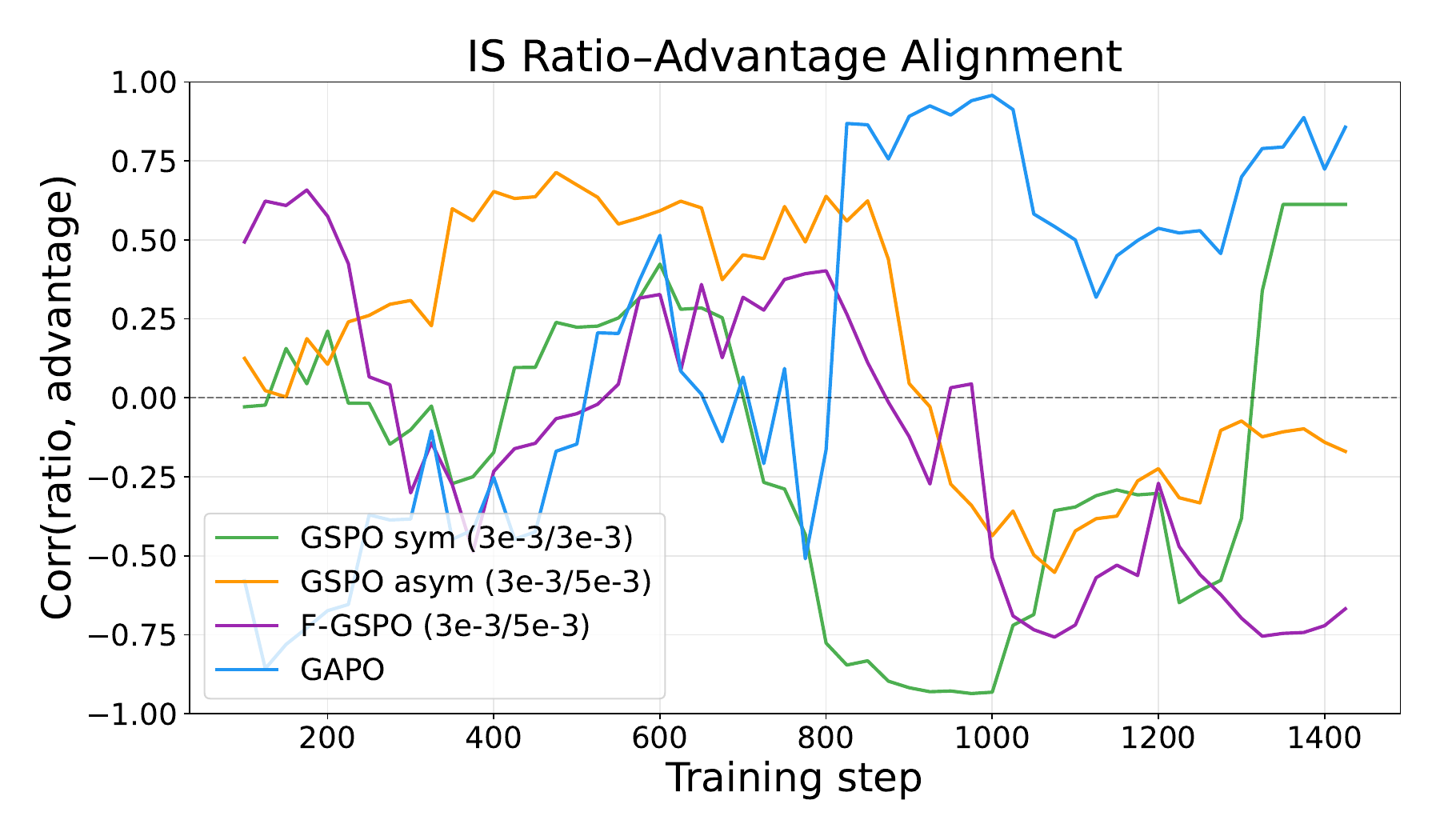}
  \caption{\textbf{IS--advantage correlation over training.} Windowed Pearson correlation between per-$c$ IS-ratio deviation and advantage ($W = 200$ steps). Positive correlation reflects the trust-region-optimal relationship $\rhoistar \propto \exp(\Ai/\lambda)$~\eqref{eq:optimal_policy}. GAPO maintains $r > 0.8$ in late training; fixed-clip baselines degrade toward negative correlation.}
  \label{fig:IS_ratio_corr}
\end{figure}
We compare GAPO against three fixed-clip baselines on Qwen2.5-Math-1.5B across clip fraction (Fig.~\ref{fig:clip_frac}), IS--advantage correlation (Fig.~\ref{fig:IS_ratio_corr}), and validation pass@1 / pass@256 (Fig.~\ref{fig:valid_pass}).

Under the reverse-KL trust region at a single prompt, the optimal
target IS ratio~\eqref{eq:optimal_policy} is proportional to
advantage (to first order). We do not directly optimise toward
this target; we instead clip proportional to advantage. Even so,
Figure~\ref{fig:IS_ratio_corr} shows GAPO retains a high
correlation between the importance-sampling ratio and advantage
throughout training, whereas fixed clipping breaks the
correlation by clipping scarce correct rollouts at the same
fraction as abundant correct ones.\\
\textbf{Step0 -- Step400}. Qwen2.5-Math-1.5B
has no reasoning capability, so the correlation
at step 0 reflects sampling noise from RL and datasampler initialization rather than RL dynamics. \\
\textbf{Step400 -- Step600}. This is a clip-free regime for scarce correct rollouts from low $c$ groups. This results in a consistent
rise in correlation across all baseline methods. Figure~\ref{fig:clip_frac} shows clipping fires
infrequently in this window, so the advantage--ratio
coupling is preserved. \\
\textbf{Step600+}. The baselines only
diverge once the clip fraction rises, at which
point GAPO sustains the correlation while fixed clipping
collapses it. 
Figure~\ref{fig:valid_pass} shows GAPO also sustains higher pass@1 and pass@256 over the same window. Appendix \ref{sec:intervention_study} provides the intervention evidence that clipping itself, rather than a common cause, drives the divergence.
Beyond aggregated metrics, Figure \ref{fig:heatmap} shows GAPO maintains problem coverage.  While baselines progressively lose solve rates on medium-difficulty AIME24 problems, GAPO retains 10.9 problems with $>5\%$ solve rate in late training compared to 6.7-8.6 for baselines. Appendix \ref{sec:appendix_diverse} shows an example of reasoning paths. 
\camera{\subsubsection{Ruling out confounds before clipping}
 Since the GSPO-asym uniform clipping baseline began firing aggressively around step 600, we took the step-600 checkpoint and switched to adaptive clipping to continue RLVR. Figure \ref{fig:ruleout_confounds_IS_adv_correlation_ckpt600_branch} in Appendix \ref{sec:intervention_study} shows  GAPO continue-finetuned (ckpt600) maintains higher adv-IS correlation and pass@k, ruling out factors before aggressive clipping (e.g. reduced advantage spectrum) as the cause of the decline in advantage-IS correlation.}  

\subsection{Benchmark results}
\label{sec:exp-benchmarks}
\definecolor{focalrow}{RGB}{232,240,254}  

\begin{table}[t!]
\centering
\small
\setlength{\tabcolsep}{2.8pt}
\renewcommand{\arraystretch}{1.15}
\resizebox{\columnwidth}{!}{%
\begin{tabular}{|l|>{\centering\arraybackslash}p{1.9cm}>{\centering\arraybackslash}p{1.9cm}>{\centering\arraybackslash}p{1.9cm}>{\centering\arraybackslash\columncolor{focalrow}}p{1.9cm}|}
\hline
\multicolumn{5}{|c|}{\textbf{Base model: DeepSeek-R1-Distill-Qwen-1.5B. Data: DeepScaleR }} \\
\multicolumn{5}{|c|}{\textbf{Pass@1 / Pass@16 (Temperature=0.6, Top-p=1, n=16, T{max}=24576)}} \\
\hline
{\textbf{In-domain}} & {\textbf{Base}} & {\textbf{GSPO}} & {\textbf{F-GSPO}} & {\textbf{GAPO}} \\
\hline
AIME24    & 28.5/60.0 & 41.3/73.3 & 40.2/73.3 & \textbf{44.0/76.7}  \\
AIME25    & 22.3/43.3 & 29.4/50.0 & 29.6/46.7 & \textbf{30.8/56.7} \\
AMC       & 71.4/\textbf{95.0} & 79.2/\textbf{95.0} & 82.2/\textbf{95.0} & \textbf{83.44/95.0} \\
MATH500   & 70.4/82.4 & 83.0/\textbf{94.7} & \textbf{85.5}/93.2 & \textbf{85.5}/93.2 \\
Minerva   & 19.0/47.4 & 26.6/50.4 & 27.9/44.3 & \textbf{29.1}/48.9 \\
Olympiad  & 41.5/60.3 & 57.8/76.2 & 59.0/72.4 & \textbf{59.4/76.2} \\
\hline
\end{tabular}}
\vspace{0.4em}
\caption{\textbf{Comparison of GAPO with GSPO baselines on DeepSeek-R1-Distill-Qwen-1.5B.} Pass@1 / Pass@16 across six in-domain math benchmarks at group size $N{=}8$. \textit{Base} is the pretrained model evaluated without RL training.  }
\label{tab:gspo_variants}
\end{table}
\begin{table}[htbp]
\centering
\scriptsize
\setlength{\tabcolsep}{5pt}
\renewcommand{\arraystretch}{1.15}
\begin{tabular}{|l|cc>{\columncolor{focalrow}}c|}
\hline
\multicolumn{4}{|c|}{\textbf{Base model: DeepSeek-R1-Distill-Qwen-1.5B. Data: DeepCoder}} \\
\multicolumn{4}{|c|}{\textbf{Pass@1 (Temperature=0.6, Top-p=1, n=2, T{max}=24576)}} \\
\hline
\textbf{In-domain} & \textbf{Base}
  & \begin{tabular}{@{}c@{}}\textbf{DeepCoder-1.5B}\\\textbf{(reproduce)}\end{tabular}
  & \begin{tabular}{@{}c@{}}\textbf{GAPO-}\\\textbf{DeepCoder-1.5B}\end{tabular} \\
\hline
LCB-v5 (8/1/24-2/1/25) & 16.9 & 22.4 & \textbf{24.8} \\
HumanEval+ & 58.3 & 68.2 & \textbf{71.7} \\
\hline
\end{tabular}
\vspace{0.4em}
\caption{\textbf{GAPO remains effective in the coding domain.} \textit{DeepCoder-1.5B} is our reproduction of the DeepCoder recipe~\citep{luo2025deepcoder} with T{max}=24576. \textit{GAPO-DeepCoder-1.5B} applies adaptive clipping on top of that recipe with $\epshimax \in [0.2, 0.28]$ (token-IS).}
\label{tab:coding_appendix}
\end{table}
\definecolor{focalrow}{RGB}{232,240,254}  

\begin{table*}[t!]
\centering
\scriptsize
\setlength{\tabcolsep}{2.8pt}
\renewcommand{\arraystretch}{1.15}
\resizebox{\textwidth}{!}{%
\begin{tabular}{|l|cc|c|cccccc|c|}
\hline
\multirow{2}{*}{\textbf{Method}} & \multirow{2}{*}{$\epslowbold$} & \multirow{2}{*}{$\epshighbold$} & \multicolumn{7}{c|}{\textbf{In-domain}} & \multicolumn{1}{c|}{\textbf{OOD}} \\
\cline{4-10}\cline{11-11}
 & & & \textbf{Avg.} & \textbf{AIME24} & \textbf{AIME25} & \textbf{AMC} & \textbf{MATH500} & \textbf{Minerva} & \textbf{Olympiad} & \textbf{IFEval} \\
\hline\hline
\multicolumn{3}{|c|}{\textbf{Qwen2.5-1.5B-Math}} & \multicolumn{8}{c|}{\textbf{Pass@1 / Pass@256\,\,(Temperature=1.0, Top-p=1, n=256, T{max}=3092)}} \\
\hline
GRPO    & \textcolor{gray!40}{0.2} & \textcolor{gray!40}{0.2} & 36.7/74.4 & 13.8/61.1 & 9.9/58.0 & 53.1/96.2 & 75.4/95.6 & 31.9/61.1 & 36.3/74.3 & 12.2 \\
F-GRPO  & \textcolor{gray!40}{0.2} & \textcolor{gray!40}{0.2} & 36.3/74.5 & 13.0/60.7 & 10.5/57.9 & 51.6/95.9 & 74.7/96.1 & 31.0/61.0 & 37.0/75.5 & 11.4 \\
DrGRPO  & \textcolor{gray!40}{0.2} & \textcolor{gray!85}{0.28} & 33.8/75.1 & 13.1/60.0 & 8.85/63.3 & 51.8/97.5 & 70.5/94.6 & 22.3/62.1 & 36.0/73.2 & 19.4 \\
\rowcolor{focalrow}
GAPO-token-IS & \textcolor{gray!40}{0.2} & \textcolor{gray!85}{0.28} & \textbf{37.6/75.6} & \textbf{14.6/63.3} & {10.2}/60.0 & \textbf{54.9/97.5} & \textbf{75.8/96.4} & \textbf{32.3/62.1} & \textbf{37.6}/74.2 & \textbf{20.9} \\
\hdashline
GSPO   & \textcolor{gray!40}{$3\ee{-3}$} & \textcolor{gray!40}{$3\ee{-3}$} & 36.8/76.2 & 15.3/60.0 & 9.28/60.0 & 53.0/100  & 75.2/97.0 & 30.9/62.9 & 37.3/77.3 & 22.1 \\
GSPO   & \textcolor{gray!40}{$3\ee{-3}$} & \textcolor{gray!85}{$5\ee{-3}$} & 37.7/74.1 & 16.2/60.0 & 9.54/56.7 & 55.3/95.0 & 75.3/95.6 & 31.0/63.6 & 37.6/73.7 & 20.1 \\
F-GSPO & \textcolor{gray!40}{$3\ee{-3}$} & \textcolor{gray!85}{$5\ee{-3}$} & 36.9/76.1 & 15.4/63.3 & 9.44/56.7 & 53.7/97.5 & 75.0/97.0 & 30.7/63.4 & 37.1/78.5 & 21.0 \\
\rowcolor{focalrow}
\textbf{GAPO} & \textcolor{gray!40}{$3\ee{-3}$} & \textcolor{gray!85}{$5\ee{-3}$} & \textbf{37.9/76.3} & \textbf{17.9/63.3} & \textbf{10.6}/56.7 & 53.4/97.5 & \textbf{76.5/97.5} & \textbf{31.2/64.1} & \textbf{38.0/78.5} & \textbf{22.2} \\
\hline
\multicolumn{3}{|c|}{\textbf{Llama3.2-3B-Instruct}} & \multicolumn{8}{c|}{\textbf{Pass@1 / Pass@256\,\,(Temperature=1.0, Top-p=1, n=256, T{max}=8192)}} \\
\hline
GRPO    & \textcolor{gray!40}{0.2} & \textcolor{gray!40}{0.2} & 23.0/59.9 & 10.7/40.7 & 0.7/21.5 & 30.5/88.2 & 55.0/90.6 & 21.8/59.0 & 19.4/59.3 & 54.1 \\
F-GRPO  & \textcolor{gray!40}{0.2} & \textcolor{gray!40}{0.2} & 23.0/63.4 & 12.1/46.1 & 1.0/29.5 & 29.8/90.6 & 54.1/92.9 & 21.0/60.1 & 20.1/61.3 & 56.4 \\
\hdashline
GSPO   & \textcolor{gray!40}{$3\ee{-3}$} & \textcolor{gray!40}{$3\ee{-3}$} & 22.7/56.5 & 13.6/36.7  & 0.4/16.7  & 27.8/86.5 & 51.5/87.8 & 20.3/57.6 & 22.3/53.9 & 54.9 \\
GSPO   & \textcolor{gray!40}{$3\ee{-3}$} & \textcolor{gray!85}{$5\ee{-3}$} & 23.0/59.2 & 14.4/46.7 & 0.7/20.0 & 27.9/87.5 & 52.0/89.2 & 20.5/55.7 & 22.4/56.1  & 54.8 \\
F-GSPO & \textcolor{gray!40}{$3\ee{-3}$} & \textcolor{gray!85}{$5\ee{-3}$} & 23.8/59.2 & 13.2/36.7 & 0.9/23.3 &  31.1/92.5 & 54.0/87.8 & 21.1/57.6 & 22.5/57.5 & 56.3 \\
\rowcolor{focalrow}
\textbf{GAPO} & \textcolor{gray!40}{$3\ee{-3}$} & \textcolor{gray!85}{$5\ee{-3}$} & \textbf{24.1/63.4} & \textbf{14.6/53.3} & \textbf{1.0}/23.3 & \textbf{31.6}/91.2 & 54.6/\textbf{93.2} & 20.3/\textbf{60.5} & 22.4/58.9 & \textbf{56.4} \\
\hline
\end{tabular}}
\vspace{0.4em}
\caption{\textbf{Comparison of GAPO with fixed-clip RLVR baselines on two base models.} Pass@1 / Pass@256 on six in-domain math benchmarks (Avg., AIME24, AIME25, AMC, MATH500, Minerva, Olympiad) and one OOD benchmark (IFEval) for Qwen2.5-1.5B-Math and Llama-3.2-3B-Instruct, trained at group size $N{=}8$. Columns $\epslowbold$ and $\epshighbold$ give the lower and upper clip thresholds (equal values denote symmetric clipping; differing values denote asymmetric clipping). Blue shaded rows mark our methods. \camera{To assess the statistical significance, 95\% confidence intervals and pairwise significance tests over 3 training seeds are reported in Table \ref{tab:ci_qwen}, \ref{tab:pairwise_sig} respectively.
}
}
\label{tab:all_models}
\end{table*}
Tables~\ref{tab:gspo_variants}, ~\ref{tab:coding_appendix}, and~\ref{tab:all_models} report
pass rates on three base models and the \camera{nine} benchmarks.\,\camera{Results are from one run per configuration, as is common in RLVR. Seeded runs on Qwen2.5-1.5B-Math appear in Table~\ref{tab:ci_qwen},~\ref{tab:pairwise_sig}.} GAPO is compared against fixed-clip RLVR
methods (GRPO, Dr.GRPO, symmetric and asymmetric GSPO) and the
focal-shaping variants (F-GRPO, F-GSPO) at matched clip widths.

\paragraph{DeepSeek-R1-Distill-Qwen-1.5B.}
On a DS-R1 distilled model with longer rollouts
($T_{\max} = 24576$), GAPO leads or matches pass@1 on all six in-domain
benchmarks (Table~\ref{tab:gspo_variants}), with the largest
gains on AIME24 ($44.0 / 76.7$, $+2.7 /\!+3.4$ over GSPO) and
AIME25 ($30.8 / 56.7$, $+1.4 / \!+6.7$ over GSPO).  \camera{Table \ref{tab:coding_appendix} shows GAPO remains effective in RLVR with code generation tasks. }

\paragraph{Qwen2.5-1.5B-Math.}
Table \ref{tab:all_models} shows GAPO achieves the best average pass@1 / pass@256 ($37.9 / 76.3$) among GSPO variants, and the highest AIME24 pass@1 ($17.9$, vs.\ $15.4$ for F-GSPO and $16.2$ for asymmetric GSPO). At the token-IS clip range, GAPO-token-IS outperforms GRPO, F-GRPO, and Dr.GRPO on average and leads pass@1 on five of the six math benchmarks, with the largest gains on AIME24 pass@1 ($14.6$) and IFEval ($20.9$ vs.\ $11.4$--$19.4$ for the baselines). \camera{These gains are significant on at least 3 of 6 benchmarks against every baseline (Table \ref{tab:ci_qwen}, \ref{tab:pairwise_sig}). }

\paragraph{Llama-3.2-3B-Instruct.}
GAPO again leads on average pass@1 / pass@256 ($24.1 / 63.4$) and improves AIME24 pass@1 over the fixed-clip GSPO
variants ($14.6$ vs.\ $10.7$--$14.4$). GAPO also achieves the best IFEval ($56.4$),
tying F-GRPO.

Across all three base models, GAPO improves pass@1 without sacrificing pass@$k$. Gains concentrate on the harder benchmarks (AIME24, +1.72 over asymmetric GSPO, significance p<0.001 in Table \ref{tab:pairwise_sig}) and shrink on easier ones, where AMC regresses against the same baseline (Table \ref{tab:all_models}, \ref{tab:pairwise_sig}). This is expected, since adaptive clipping redistributes gradient toward scarce correct rollouts, which are only scarce on problems the model rarely solves.

\subsection{Ablation Study}
\begin{figure}[t]
    \includegraphics[width=\linewidth]{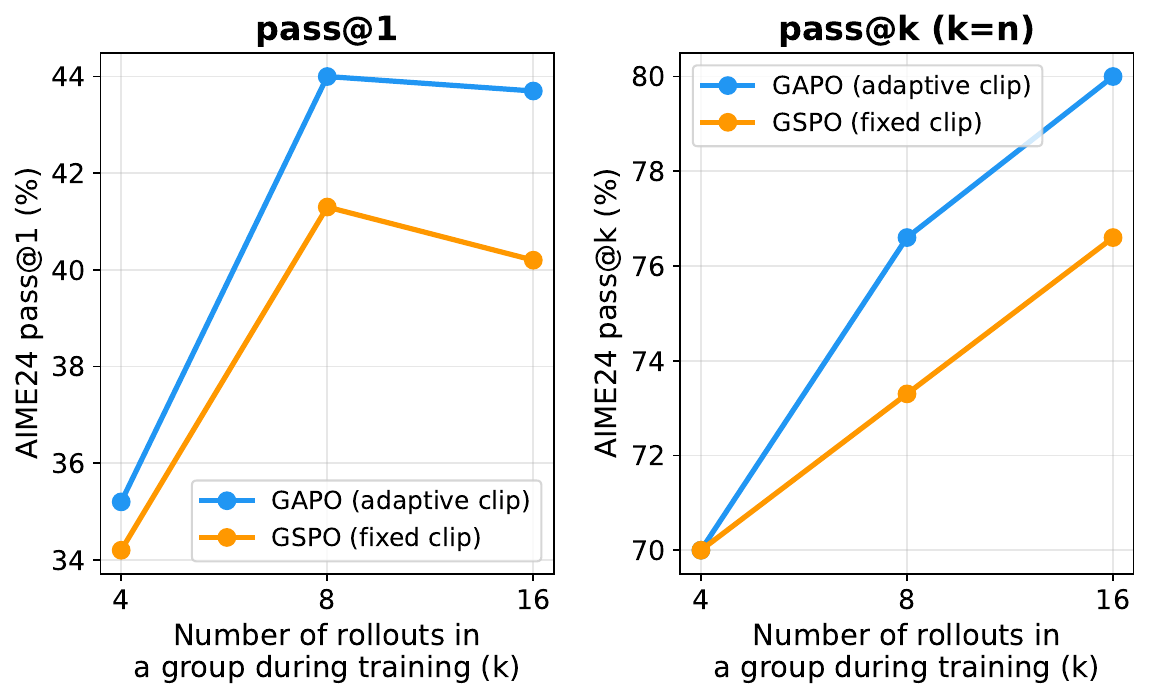} \hfill
    \caption{Ablation on the number of rollouts per group $k$
used during training. Scaling from $k{=}4$ to $k{=}8$ yields
clear improvement in both pass@1 and pass@256 on AIME24 eval, and widens
the gap between fixed clipping (GSPO) and adaptive clipping
(GAPO).    }
    \label{fig:ablationrollouts}
\end{figure}

We ablate \camera{three} key algorithm choices for GAPO. 
\textbf{Seq-IS vs token-IS.}\,In
Table~\ref{tab:all_models}, the more principled sequence-level
adaptive IS clipping (AIME24 pass@1 $17.9\%$, math average
pass@1 $37.9\%$) outperforms token-level adaptive IS clipping
(AIME24 pass@1 $14.6\%$, math average  $37.6\%$).\\
\textbf{Group size $k$.}\, $k$ controls how finely advantages can be differentiated within a group, and hence the resolution of the per-prompt clip threshold. Increasing $k$ from $4$ to $8$   further widens the performance gap between GAPO and GSPO (Figure~\ref{fig:ablationrollouts}). On the math training data, increasing $k$ beyond $8$ shows no \camera{noticeable} gain, but the training was more stable.
\\
\camera{\textbf{Adaptive upper clipping bound} $\epshimax$. Across the range in Table \ref{tab:sweepcompare}, GSPO with fixed $\epshigh=\epshimax$ is worse than GAPO on AIME24. Because adaptive clipping constrains abundant correct rollouts to save the trust-region budget, slightly relaxing the value $\epshimax=5\ee{-3}$, which was tuned for fixed-clipping, to $5.25\ee{-3}$ yields a small improvement.} 

\section{Related Work}
\label{sec:related_work}
\subsection{LLM Post-training}
\textbf{SFT.} s1~\citep{muennighoff2025s1_simpletesttime} shows that SFT on only 1K high-quality DeepSeek-R1 \citep{guo2025deepseekR1} traces can surpass o1-preview \citep{jaech2024openaio111}. SSFT~\citep{jia2026training} uses a matching loss to distill from multiple teachers to capture diverse reasoning modes. On-policy distillation~\citep{agarwal2024onpolicy} leverages the access to teacher logits to perform token-level distribution matching between the student and the teacher.\\
\textbf{RLVR.} \citet{khatri2025art} surveys group-relative RLVR methods.\,\,For token loss aggregation, Dr.GRPO~\citep{liu2025understanding_r1zero} normalizes by a constant and DAPO~\citep{yu2025dapo_NEURIPS2025_a4277440} by total batch tokens, both to avoid length bias. For advantage normalization, Dr.GRPO removes the within-group std rescaling to preserve unbiasedness, while REINFORCE++~\citep{hu2025reinforce++} substitutes batch-level std. All use token-level IS with fixed clipping, with DAPO widening the upper threshold asymmetrically. GSPO~\citep{zheng2025group} instead uses sequence-level IS and clipping. CISPO~\citep{chen2025minimax} clips the IS weight directly while preserving every token's gradient, a weighted REINFORCE. DAPO's clip-higher and CISPO's gradient preservation share our motivation of retaining exploration tokens. We arrive at the same goal through an explicit closed-form derivation: the per-prompt reverse-KL trust-region-optimal IS ratio suggests a clip threshold proportional to advantage.  Other exploration-oriented approaches scale the number of rollouts~\citep{hu2025brorlScalingRLviaBroadenedExploration}, introduce separate fixed clipping thresholds for correct and incorrect rollouts~\citep{karaman2026dispo}, or only optimize policy on exploratory forking tokens \citep{wang2025beyond_80_20_rule, wang2025emergenthierarchical}.

\subsection{Advantage and Reward Shaping for Exploration in RLVR}
\citet{yue2025doesRLreallyincentivize} found that pass@$k$ of RLVR-trained LLMs score below the base model at large $k$, motivating a line of work that directly optimizes pass@$k$~\citep{chen2025passktrainingbytedance, walder2025passkoptimization_NEURIPS2025_df8a1a63, thrampoulidis2025advantageasrewardshaping}. However, pass@1, the deployment metric, has been observed to degrade under this training~\citep{walder2025passkoptimization_NEURIPS2025_df8a1a63, barakat2026passkcandegradepassone}. Rather than directly optimizing pass@$k$, others propose dedicated advantage-shaping schemes to encourage exploration. F-GRPO~\citep{plyusov2026f} and \citet{he-etal-2025-rewarding} apply advantage shaping based on within-group success count; \citet{zhou2025daroDfficultyAwareReweighting} reweights the loss across groups by success rate; and \citet{gai2025differentialSmoothingSharpeningAndImproves} adds a differential-smoothing reward based on per-rollout correctness. In contrast, GAPO still optimizes the pass@1 objective, encouraging exploration by maintaining high correlation between the empirical IS ratio and advantage, which preserves gradients on rare correct rollouts relative to redundant ones, according to the target IS ratio for the reverse KL trust region formulation.

\section{Conclusion}
\label{sec:conclusion}
We provide a simple plug-in adaptive clipping method, Group Adaptive Policy Optimization (GAPO), motivated by the reverse-KL trust-region-optimal IS ratio at a single prompt. Under this trust region, the optimal IS ratio scales exponentially with advantage, suggesting that the per-rollout clip threshold should scale with advantage rather than be applied uniformly. In RLVR, binary rewards reduce this to a closed-form schedule indexed by the group correctness count $c$, requiring no new hyperparameters beyond the existing clip range. Across Qwen2.5-Math-1.5B, Llama-3.2-3B-Instruct, and DeepSeek-R1-Distill-Qwen-1.5B on mathematical reasoning \camera{and competitive coding benchmarks}, GAPO consistently improves pass@1 while retaining pass@$k$, and maintains a high IS--advantage correlation throughout training where fixed clipping collapses it. Unlike reward-shaping and advantage-shaping approaches for exploration, GAPO still optimizes the deployment metric pass@1 directly while implicitly encouraging exploration through optimally allocating trust-region constraint.

\section*{Limitations}
\label{sec:limitations}
\textbf{Fixed $\epslow$ across rollouts.} We adapt only the upper clip threshold $\epshigh$ based on $A_i$, which governs positive-advantage rollouts. The lower clip $\epslow$ for negative-advantage (incorrect) rollouts is held fixed throughout. \camera{Because $\epslow$ acts only on incorrect rollouts, fixing it leaves the positive learning signal GAPO targets untouched. In future work, we plan to study how jointly adapting the headroom for penalizing incorrect rollouts based on task difficulty indirectly affects the model's exploration behavior.}\\
\textbf{Single prompt trust-region analysis.} 
 Our derivation of the per-rollout optimal IS ratio $\rhoistar$ is based on the reverse-KL trust region at a single prompt $x$ (Section~\ref{sec:gapo-derivation}). \camera{Since \eqref{eq:reverse_constrained} constrains the KL averaged over the batch, the multiplier $\lambda$ is in principle shared across prompts; we treat it as a global constant, which is what makes \eqref{eq:gapo} depend on $c$ alone. RLVR runs always have $>1$ prompts or tasks in a mini-batch, so the gap is real, though} the observed IS-advantage correlations (Figure~\ref{fig:IS_ratio_corr}) suggest the per-prompt heuristic still provides useful empirical guidance.

%
\section*{Ethical Considerations}
\label{sec:ethics}

\paragraph{Scope and broader impact.} GAPO is a methodological contribution to RLVR: it adapts the per-rollout clipping threshold without introducing new model capabilities, training data, or evaluation protocols beyond those standard in the field. The intended impact is to improve the training efficiency and final performance of LLMs on mathematical reasoning with applications in education.

\paragraph{Data and models.} All training data (DeepScaleR, DeepCoder-preview-dataset) and evaluation benchmarks (AIME24/25, MATH500, AMC, Olympiad, Minerva, LCB-v5, HumanEval+, IFEval) are publicly available. The base models (Qwen2.5-Math-1.5B, Llama-3.2-3B-Instruct, DeepSeek-R1-Distill-Qwen-1.5B) are openly released. No human subjects, annotators, or proprietary data were involved.

\paragraph{Dual-use considerations.} As with any improvement to LLM training, GAPO may indirectly contribute to producing more capable reasoning systems, which carry the general dual-use considerations common to language modeling research. GAPO does not specifically lower the cost of generating harmful content, nor does it target capabilities relevant to harmful applications; we do not anticipate risks beyond those inherent to RLVR training in general.

\section*{Acknowledgements}
We used Claude Opus 4.7 \citep{claude_41_opus} for occasional grammar checks on individual sentences, which were entered manually through the chat interface. No text, code, experimental results, or figures were LLM-generated. All hypotheses, technical content, and conclusions are our own, and we verified them ourselves.
%

\bibliography{custom}
\appendix

\section{Training Hyperparameters}
\begin{table}[!h]
\centering
\footnotesize
\setlength{\tabcolsep}{4pt}
\begin{tabular}{@{}l>{\raggedright\arraybackslash}p{0.50\columnwidth}@{}}
\toprule
\textbf{Parameter} & \textbf{Value} \\
\midrule
\multicolumn{2}{c}{\textit{Optimization}} \\
Trainer Precision & bfloat16  \\
Optimizer & AdamW \citep{loshchilov2017decoupled} \\
Learning rate & $1 \times 10^{-6}$ \\
LR schedule & Constant \\
Weight decay & 0.01 \\
Gradient clip & 1.0 \\
KL coefficient $\lambda$ & $0$ \\
Entropy coefficient & $0$ \\
Reward & Verifiable (0/1) \\
Training steps & 10 epochs \\
\addlinespace
\multicolumn{2}{c}{\textit{RL Settings}} \\
Inference Precision & bfloat16  \\
Training dataset & DeepScaleR \citep{tan2026deepscaler}, 39,202 samples \\
Rollout batch (prompts) & 256 \\
Mini-batch (prompts) & 64 \\
Rollouts per prompt $k$ & 8 or 16 \\
Token-IS clip $(\epslow, \epshigh)$ & $(0.2,\ 0.28)$ \\
Seq-IS clip $(\epslow, \epshigh)$ & $(3e{-}3,\ 5e{-}3)$\\
& $(7e{-}5,\ 3e{-}4)$ distilled base \\
\addlinespace
\multicolumn{2}{c}{\textit{Generation}} \\
Top-$p$ & 1.0 \\
Max prompt length & 2048 \\
\addlinespace
\hdashline
\addlinespace[2pt]
\multicolumn{2}{c}{\scriptsize{\textit{Qwen2.5-1.5B-Math/\,Llama-3.2-3B-Instruct/\,DS-R1-Distill-Qwen-1.5B}}} \\
\addlinespace[2pt]
\hdashline
\addlinespace
Max response length & 3092 / 8192 / 24576 \\
Temperature & 1.0 / 1.0 / 1.0 \\
\bottomrule
\end{tabular}
\caption{\textbf{Training hyperparameters used in all math RL experiments.}  We use 8xH200 GPUs with the verl \citep{sheng2025hybridflow} framework for both training and evaluation. Training takes about 4--5 days for DS-R1 distilled models and 1 day for Qwen2.5-Math-1.5B and Llama models. These are used for Table~\ref{tab:gspo_variants}, \ref{tab:all_models}, Figure \ref{fig:valid_pass}. For all of our experiments, we use DP=8, TP=1, SP=1 with FSDP.}
\label{tab:hyperparams}
\end{table}

\begin{table}[!ht]
\centering
\footnotesize
\setlength{\tabcolsep}{4pt}
\begin{tabular}{@{}l>{\raggedright\arraybackslash}p{0.50\columnwidth}@{}}
\toprule
\textbf{Parameter} & \textbf{Value} \\
\midrule
\multicolumn{2}{c}{\textit{Optimization hyperparameters are the same as Table \ref{tab:hyperparams}}} \\
\addlinespace
\multicolumn{2}{c}{\textit{RL Settings}} \\
Inference Precision & bfloat16  \\
Training dataset & DeepCoder-Preview-Dataset \citep{luo2025deepcoder}, 24,269 samples consisting of 16{,}238  SYNTHETIC-1~\citep{2025synthetic1}, 7{,}432 from TACO~\citep{likaixin2024taco-verified}, and 599 from LiveCodeBench (2023/5/1-2024/7/31) \citep{jain2025livecodebench}. \\
Rollout batch (prompts) & 256 \\
Mini-batch (prompts) & 64 \\
Rollouts per prompt $k$ & 8 or 16 \\
Token-IS clip $(\epslow, \epshigh)$ & $(0.2,\ 0.28)$ \\
\addlinespace
\multicolumn{2}{c}{\textit{Generation parameters are the same as Table \ref{tab:hyperparams}} } \\
\multicolumn{2}{c}{ \textit{under DS-R1-Distill-Qwen-1.5B}}\\
\bottomrule
\end{tabular}
\caption{\textbf{Hyperparams in coding RL runs}  for  Table \ref{tab:coding_appendix}.}
\label{tab:hyperparams_coding}
\end{table}

\section{Intervention study}
\label{sec:intervention_study}
\begin{figure}[h]
\centering
\includegraphics[width=0.7\linewidth]{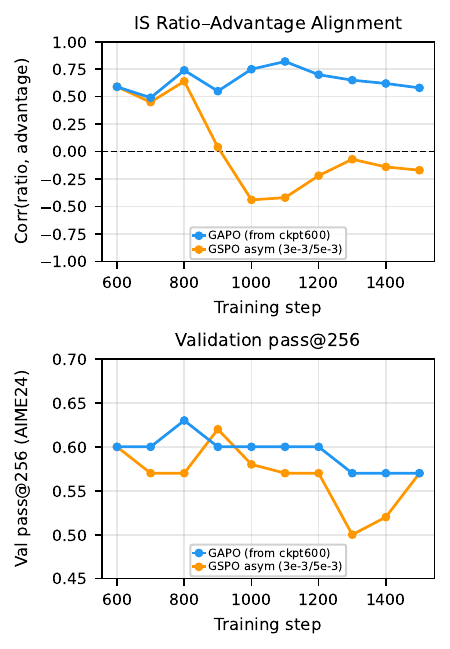}
 \caption{\camera{\textbf{Clipping, not an earlier factor, drives the correlation decline.} Both branches start from GSPO-asym step 600. Adaptive clipping (GAPO) sustains adv-IS correlation and Pass@256, uniform clipping (GSPO-asym) declines. (Training data: DeepScaleR)}}
\label{fig:ruleout_confounds_IS_adv_correlation_ckpt600_branch}
\end{figure}

\camera{To isolate uniform clipping as the cause of the IS-advantage correlation decline, we branch from the GSPO-asym step-600 ckpt, where uniform clipping starts firing aggressively, and switch to adaptive clipping. The GAPO branch sustains advantage-IS correlation and Pass@256 while the uniform branch declines, ruling out earlier factors (e.g. reduced advantage spectrum).  (Figure~\ref{fig:ruleout_confounds_IS_adv_correlation_ckpt600_branch}) }
\clearpage

\onecolumn
\section{Robustness Analysis}
We stress-test the main-table results (Table~\ref{tab:all_models}) along two axes: statistical significance across training seeds, and sensitivity to the clip-width hyperparameter. Table \ref{tab:all_models} follows the common practice in RLVR. Seeded runs cover the Qwen2.5-1.5B-Math setting, since compute constraints made 3-seed runs infeasible for all configurations. So Table~\ref{tab:ci_qwen} means differ slightly from the single-seed entries in Table~\ref{tab:all_models}.  
\subsection{Significance Tests}
\camera{Table~\ref{tab:ci_qwen} reports 3-seed means with 95\% confidence intervals, and Table~\ref{tab:pairwise_sig} the paired per-seed differences with Welch's $t$-test significance. GAPO's Pass@1 gains are significant on at least 3 of 6 benchmarks against every baseline, and strongest on harder reasoning benchmarks like AIME24.
}
\definecolor{focalrow}{RGB}{232,240,254}  
\begin{table*}[htbp]
\centering
\scriptsize
\setlength{\tabcolsep}{2.8pt}
\renewcommand{\arraystretch}{1.15}
\resizebox{\textwidth}{!}{%
\begin{tabular}{|l|cc|cccccc|}
\hline
\multirow{2}{*}{\textbf{Method}} & \multirow{2}{*}{$\epslowbold$} & \multirow{2}{*}{$\epshighbold$} & \multicolumn{6}{c|}{\textbf{Qwen2.5-1.5B-Math\,\,\,Pass@1\,\,(Temperature=1.0, Top-p=1, n=256, T{max}=3092)}} \\
\cline{4-9}
 & & & \textbf{AIME24} & \textbf{AIME25} & \textbf{AMC} & \textbf{MATH500} & \textbf{Minerva} & \textbf{Olympiad} \\
\hline
GRPO          & \textcolor{gray!40}{0.2} & \textcolor{gray!40}{0.2}  & 13.82$\pm$0.81 & 10.07$\pm$0.79 & 53.16$\pm$1.26 & 75.61$\pm$0.99 & 31.33$\pm$0.94 & 36.90$\pm$1.31 \\
F-GRPO        & \textcolor{gray!40}{0.2} & \textcolor{gray!40}{0.2}  & 12.95$\pm$0.75 & 10.44$\pm$0.33 & 51.53$\pm$1.69 & 74.64$\pm$0.64 & 30.99$\pm$1.18 & 36.95$\pm$1.52 \\
DrGRPO        & \textcolor{gray!40}{0.2} & \textcolor{gray!85}{0.28} & 13.11$\pm$0.69 & 8.87$\pm$0.77 & 51.65$\pm$1.33 & 70.48$\pm$1.01 & 22.43$\pm$1.25 & 35.92$\pm$1.39 \\
\rowcolor{focalrow}
\textbf{GAPO-token-IS} & \textcolor{gray!40}{0.2} & \textcolor{gray!85}{0.28} & \textbf{14.61$\pm$0.83} & 10.19$\pm$0.48 & \textbf{54.87$\pm$0.65} & 75.69$\pm$1.03 & \textbf{32.34$\pm$0.97} & 37.53$\pm$1.08 \\
\hdashline
GSPO          & \textcolor{gray!40}{$3\ee{-3}$} & \textcolor{gray!40}{$3\ee{-3}$} & 15.28$\pm$0.74 & 9.28$\pm$0.68 & 53.12$\pm$0.92 & 75.20$\pm$0.60 & 31.05$\pm$1.10 & 37.28$\pm$0.65 \\
GSPO          & \textcolor{gray!40}{$3\ee{-3}$} & \textcolor{gray!85}{$5\ee{-3}$} & 16.24$\pm$0.61 & 9.56$\pm$0.70 & 55.23$\pm$1.19 & 75.32$\pm$0.52 & 31.07$\pm$1.42 & 37.55$\pm$1.11 \\
F-GSPO        & \textcolor{gray!40}{$3\ee{-3}$} & \textcolor{gray!85}{$5\ee{-3}$} & 15.39$\pm$0.51 & 9.34$\pm$0.77 & 53.73$\pm$1.17 & 75.08$\pm$1.05 & 30.63$\pm$1.35 & 37.16$\pm$1.41 \\
\rowcolor{focalrow}
\textbf{GAPO} & \textcolor{gray!40}{$3\ee{-3}$} & \textcolor{gray!85}{$5\ee{-3}$} & \textbf{17.96$\pm$0.48} & \textbf{10.56$\pm$0.51} & 53.37$\pm$0.83 & \textbf{76.59$\pm$1.34} & 31.70$\pm$1.60 & \textbf{38.03$\pm$0.58} \\
\hline
\end{tabular}}
\vspace{0.4em}
\caption{\textbf{95\% confidence intervals for comparing GAPO with fixed-clip RLVR baselines on Qwen2.5-1.5B-Math.} Each entry is the mean over 3 training seeds $\pm$ a 95\% confidence interval (Student-$t$, $\mathrm{df}{=}2$).  Recap on the settings: $\epslowbold$ and $\epshighbold$ are the lower and upper clipping thresholds: equal values denote symmetric clipping, differing values asymmetric clipping. The upper block uses token-level importance sampling and the lower block sequence-level (GSPO-style) importance sampling. Within each block, our method (blue) shares the importance-sampling scheme and the clip range of its baselines, so the only difference is that $\epshighbold$ is adapted per group rather than held fixed.}
\label{tab:ci_qwen}
\end{table*}
\begin{table*}[htbp]
\centering
\small
\setlength{\tabcolsep}{5pt}
\renewcommand{\arraystretch}{1.15}
\begin{tabular}{|l|cccccc|}
\hline
\textbf{Comparison ($\Delta$ = ours $-$ baseline)} & \textbf{AIME24} & \textbf{AIME25} & \textbf{AMC} & \textbf{MATH500} & \textbf{Minerva} & \textbf{Olympiad} \\
\hline
\rowcolor{focalrow}
\multicolumn{7}{|l|}{\textbf{GAPO-token-IS} (clip $0.2/0.28$) vs.\ baselines:} \\
\hline
vs.\ GRPO   & $+0.79^{*}$   & $+0.12$      & $+1.71^{*}$   & $+0.08$       & $+1.01^{*}$   & $+0.63$ \\
vs.\ F-GRPO & $+1.66^{**}$  & $-0.25$      & $+3.34^{**}$  & $+1.05^{*}$   & $+1.35^{*}$   & $+0.58$ \\
vs.\ DrGRPO & $+1.50^{**}$  & $+1.32^{**}$ & $+3.22^{**}$  & $+5.21^{***}$ & $+9.91^{***}$ & $+1.61^{*}$ \\
\hline
\rowcolor{focalrow}
\multicolumn{7}{|l|}{\textbf{GAPO} (clip $3\ee{-3}/5\ee{-3}$) vs.\ baselines:} \\
\hline
vs.\ GSPO $3/3$ & $+2.68^{***}$ & $+1.28^{**}$ & $+0.25$      & $+1.39^{*}$ & $+0.65$ & $+0.75^{*}$ \\
vs.\ GSPO $3/5$ & $+1.72^{***}$ & $+1.00^{**}$ & $-1.86^{**}$ & $+1.27^{*}$ & $+0.63$ & $+0.48$ \\
vs.\ F-GSPO     & $+2.57^{***}$ & $+1.22^{*}$  & $-0.36$      & $+1.51^{*}$ & $+1.07$ & $+0.87$ \\
\hline
\end{tabular}
\vspace{0.4em}
\caption{\textbf{Pairwise significance of Pass@1 gains.} Each entry is the mean difference $\Delta$ on Pass@1 over 3 training seeds, computed from the per-seed runs summarized in Table~\ref{tab:ci_qwen}. Significance is from a Welch's $t$-test: $^{*}$ denotes $p<0.05$, $^{**}$ denotes $p<0.01$, and $^{***}$ denotes $p<0.001$; unmarked entries are not significant. GAPO-token-IS is compared against the token-IS baselines, GAPO against the GSPO baselines.  The gains are statistically significant on at least 3 out of 6 benchmarks for every baseline comparison, and the improvement is noticeable for harder benchmarks like AIME24. There's only one meaningful regression compared to GSPO 3/5 on AMC.}
\label{tab:pairwise_sig}
\end{table*}
\clearpage

\subsection{Hyperparameter Sweeps}
\camera{Table~\ref{tab:sweepcompare} sweeps the maximum upper clip $\epshimax$: GAPO's adaptive $\epshigh$ beats uniform clipping at every level tested.}
\begin{table*}[!h]
\centering
\small
\setlength{\tabcolsep}{3pt}
\renewcommand{\arraystretch}{1.15}
\begin{tabular}{|l|l|ccccc|}
\hline
\multicolumn{7}{|c|}{\textbf{AIME24 \quad Pass@1 \quad Qwen2.5-1.5B-Math \quad (n=256, T{max}=3092)}} \\
\hline
& \textbf{Clip-high ($\epsilon_{\mathrm{hi}}$)} adaptive range
& $[3\ee{-3}]$ & $[5\ee{-3}]$ & $[5.25\ee{-3}]$ & $[7\ee{-3}]$  &\\
\multirow{-2}{*}{\textbf{GSPO asym (uniform)}} & Pass@1 & 15.28 & \textbf{16.24} & 15.98 & 15.91 & \\
\hline
\rowcolor{focalrow}
 & \textbf{Clip-high ($\epsilon_{\mathrm{hi}}$)} adaptive range & $[3\ee{-3},5\ee{-3}]$ & $[3\ee{-3},5\ee{-3}]$ & $[3\ee{-3},5.25\ee{-3}]$ & $[3\ee{-3},7\ee{-3}]$ & \\
\rowcolor{focalrow}
\multirow{-2}{*}{\textbf{GAPO (ours)}} & Pass@1 & \textbf{17.96} & \textbf{17.96} & \textbf{18.20} & \textbf{17.38} & \\
\hline
\textbf{Improvement} & ours/baseline $-$ 1 & $+17.54\%$ & $+10.59\%$ & $+13.89\%$ & $+9.24\%$ &  \\
\hline
\end{tabular}
\vspace{0.4em}
\caption{\textbf{GAPO's adaptive $\epsilon_{\mathrm{hi}}$ beats uniform clipping at every level tested.} AIME24 Pass@1 on Qwen2.5-1.5B-Math. Each column pairs a uniform fixed clipping GSPO baseline with the GAPO run it is compared against; GAPO's adaptive range is listed in its own row. $\epslow$ is fixed at $3\ee{-3}$. }
\label{tab:sweepcompare}
\end{table*}

\section{Per-Problem Solve Rate Analysis}
\begin{figure*}[!h]
    \includegraphics[width=1\linewidth]{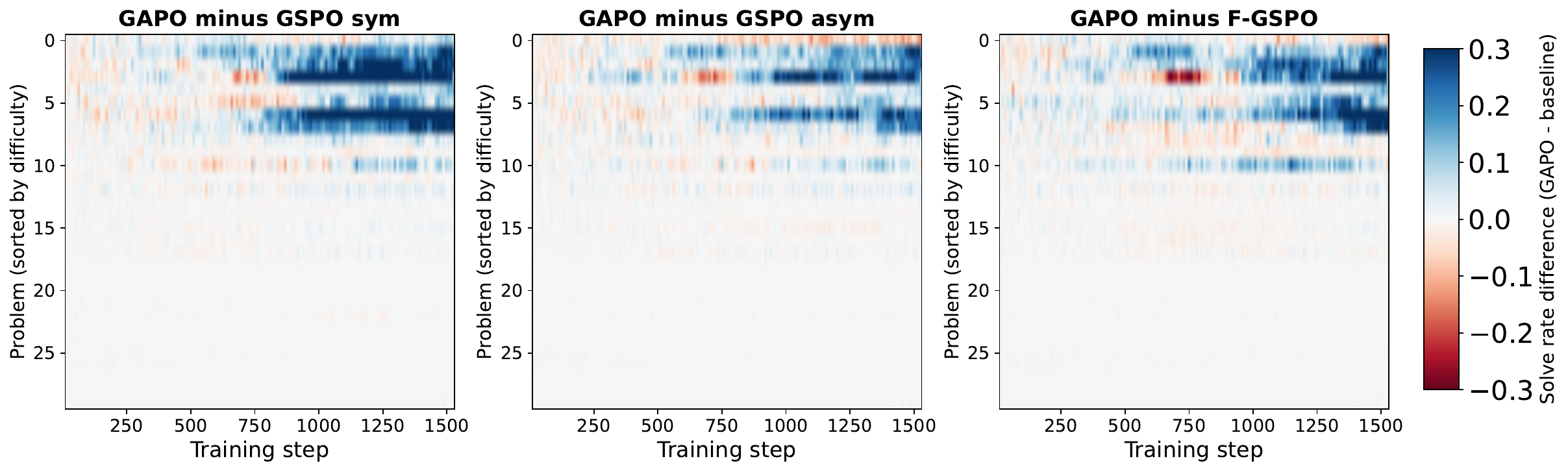}
\caption{\textbf{Per-problem solve rate difference (GAPO minus baseline) on AIME24 validation over training for Qwen2.5-Math-1.5B} (256 samples per problem, evaluated every 10 steps). Problems are sorted by difficulty (easiest at top). Blue indicates higher solve rate for GAPO; red indicates higher solve rate for the baseline. Results are averaged across all seeds per method. GAPO's advantage is concentrated on medium-difficulty problems (rows 5--12) and grows in late training as baselines progressively lose solve rate on these frontier problems. Averaged over steps 1200--1530, GAPO maintains 10.9 problems with $>5\%$ solve rate compared to 8.6 (F-GSPO), 8.5 (GSPO asym), and 6.7 (GSPO sym).}    
\label{fig:heatmap}
\end{figure*}
\clearpage

\section{Diverse Correct Solutions Found by GAPO}
\label{sec:appendix_diverse}
\begin{figure}[!h]
\small
\begin{problembox}{Problem (AIME24, Chip Placement)}
There is a collection of $25$ indistinguishable white chips and $25$ indistinguishable black chips. Find the number of ways to place some of these chips in the $25$ unit cells of a $5\times5$ grid such that:
\begin{enumerate}
\item[(i)] each cell contains at most one chip;
\item[(ii)] all chips in the same row and all chips in the same column have the same colour;
\item[(iii)] any additional chip placed on the grid would violate one or more of the previous two conditions.
\end{enumerate}
\end{problembox}
\vspace{0.6em}
\begin{thinkbox}{\texttt{<Generation 1>} --- Color the row/columns, add chips by deduction}
\texttt{<|im\_start|>assistant}\\
\texttt{<|im\_start|>}\\
``each cell at most one chip, all chips in same row same color, same column same color'' \dots\ \dots\
``rows and columns alternate in color'' \dots\ \dots\
``mixing: choose which rows are white and which are black; similarly for columns'' \dots\ \dots\
$2^{5}-2 = 30$ ways to choose rows (excluding all-white and all-black) \dots\ \dots\
$2^{5}-2 = 30$ ways for columns \dots\ \dots\
mixed configurations: $30\times 30 = 900$ \dots\ monochrome: $2$ \dots\
\textcolor{commentred}{(\textit{Comment:} each row/column picks white or black independently; cells where row and column colors agree get a chip, the rest stay empty. Excluding the trivial all-white and all-black choices gives $30\times 30=900$ mixed placements; add the $2$ monochrome grids: $902$.)}\\
\texttt{<|im\_start|>answer}\\
Answer: \dots\ \fbox{902}
\end{thinkbox}
\vspace{0.6em}
\newpage
\begin{thinkbox}{\texttt{<Generation 2>} --- Partition the grid into color blocks}
\texttt{<|im\_start|>assistant}\\
\texttt{<|im\_start|>}\\
``the grid is divided into two rectangles of the same color'' \dots\ \dots\
``all $25$ cells same color: $2$ configurations'' \dots\ \dots\
``choose a subset of rows and a subset of columns to form the two regions'' \dots\ \dots\
$2^{5}-2$ ways to divide the rows (any non-empty proper subset) \dots\
$2^{5}-2$ ways to divide the columns \dots\
$(2^{5}-2)\times(2^{5}-2) = 30 \times 30 = 900$ partitions \dots\
total $2 + 900 = 902$ \dots\
\textcolor{commentred}{(\textit{Comment:} each valid chip placement corresponds to a partition of the grid into sub-grids; counting non-trivial row/column splits gives $30\times 30$, and the two trivial monochrome cases add $2$.)}\\
\texttt{<|im\_start|>answer}\\
Answer: \dots\ \fbox{902}
\end{thinkbox}
\vspace{0.6em}
\begin{thinkbox}{\texttt{<Generation 3>} --- Alternating patterns}
\texttt{<|im\_start|>assistant}\\
\texttt{<|im\_start|>}\\
``no two chips of different colors are adjacent'' \dots\ \dots\
``each row and each column must be uniform in color'' \dots\ \dots\
``the grid is bipartite in terms of color'' \dots\ \dots\
explicit patterns enumerated: rows $\{1,3,5\}$ white, rows $\{2,4\}$ black; or vice versa; \dots\
``choose any subset of the $5$ rows to be all white, excluding all-W and all-B: $2^{5}-2 = 30$'' \dots\
``similarly $30$ choices for columns'' \dots\
$30\times 30 = 900$ \dots\ plus $2$ monochrome \dots\
\textcolor{commentred}{(\textit{Comment:} start from a specific pattern, e.g., rows $\{1,3,5\}$ white and rows $\{2,4\}$ black, recognise that any way of splitting the 5 rows into white and black works as long as both colors appear, and likewise for the columns; this generalizes to $30\times 30$ mixed patterns, plus the $2$ single-color grids.))}\\
\texttt{<|im\_start|>answer}\\
Answer: \dots\ \fbox{902}
\end{thinkbox}
\vspace{0.6em}
\noindent\begin{minipage}{\textwidth}
\captionsetup{hypcap=false}%
\captionof{figure}{\textbf{Diversity preservation on a single AIME24 problem.} The baseline GSPO produces only $3$ correct generations on this problem, while GAPO produces $28$ and, near peak validation pass@1, samples three distinct correct reasoning paths. All three converge to the same answer, but enter the problem from different conceptual angles. This suggests that GAPO retains multiple correct modes where the baseline has effectively collapsed.}
\label{fig:diverse_8dc38065}
\end{minipage}
\end{figure}

\end{document}